\documentclass[runningheads]{llncs}
\usepackage{graphicx}
\usepackage{tikz}
\usepackage{comment}
\usepackage{amsmath,amssymb} % define this before the line numbering.
\usepackage{color}
\usepackage[accsupp]{axessibility}  % Improves PDF readability for those with disabilities.

\usepackage{epsfig}
\usepackage[normalem]{ulem}
\usepackage{sidecap}
\usepackage{tabularx} % Automatically calculates column widths with text wrapping
\usepackage{algorithm,algorithmicx,algpseudocode}
\usepackage{subfigure}
\usepackage{adjustbox}
\usepackage{multirow}

\usepackage{hyperref}
\newcommand{\orcid}[1]{\href{https://orcid.org/#1}{\includegraphics[width=8pt]{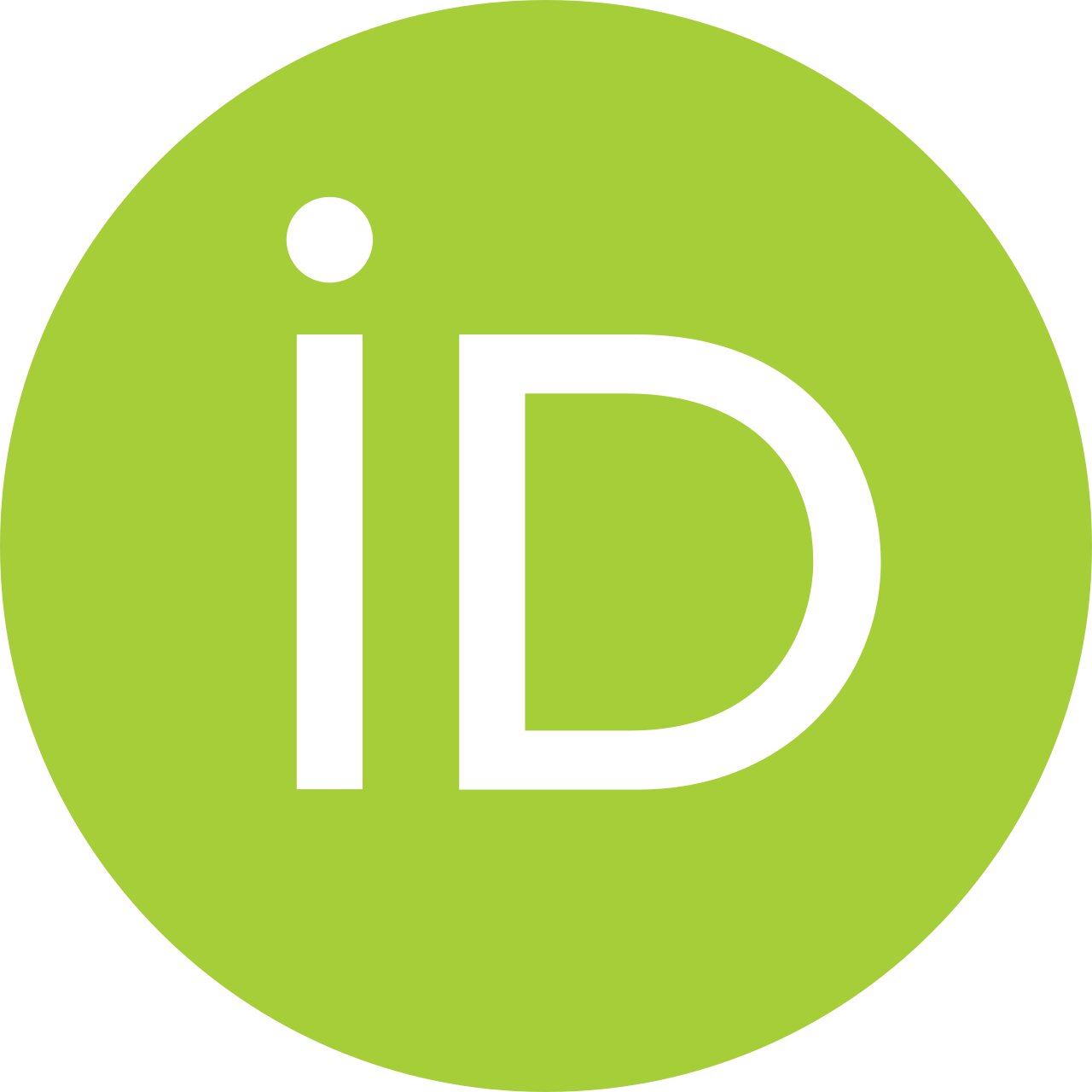}}}

\def\etal{\textit{et al.}}
\usepackage{sidecap, caption}
\usepackage{wrapfig}
\usepackage{pifont}

\begin{document}
% \linenumbers
\pagestyle{headings}
\mainmatter
\def\ECCVSubNumber{3641}  % Insert your submission number here

\title{Few-shot Class-incremental Learning for\\3D Point Cloud Objects} % Replace with your title

\begin{comment}
\titlerunning{ECCV-22 submission ID \ECCVSubNumber} 
\authorrunning{ECCV-22 submission ID \ECCVSubNumber} 
\author{Anonymous ECCV submission}
\institute{Paper ID \ECCVSubNumber}
\end{comment}
%******************

%\begin{comment}
\titlerunning{FSCIL for 3D Point Cloud Objects}
\author{Townim Chowdhury\inst{1}\orcid{0000-0003-1780-6046} \and
Ali Cheraghian\inst{2,3}\orcid{0000-0002-3324-7849} \and
Sameera Ramasinghe\inst{4}\orcid{0000-0002-3200-9291} \and\\
Sahar Ahmadi\inst{5}\orcid{0000-0002-2161-7005} \and
Morteza Saberi\inst{6}\orcid{0000-0002-5168-2078} \and
Shafin Rahman\inst{1}\thanks{Corresponding author}\orcid{0000-0001-7169-0318}
}
\authorrunning{T. Chowdhury et al.}
\institute{Dept. of Electrical and Computer Engineering, North South University, Bangladesh \and
School of Engineering, Australian National University, Australia \and
Data61, Commonwealth Scientific and Industrial Research Organisation, Australia
\and
Australian Institute for Machine Learning, University of Adelaide, Australia 
\and
Business school, The University of New South Wales,  Australia
\and
School of Computer Science and DSI, University of Technology Sydney, Australia
\\
% Corresponding Author Email: \email{\small\texttt{shafin.rahman@northsouth.edu}}
\email{\small\texttt{\{townim.faisal, shafin.rahman\}@northsouth.edu,\\ali.cheraghian@anu.edu.au,  sameera.ramasinghe@adelaide.edu.au,\\sahar.ahmadi@unsw.edu.au, morteza.saberi@uts.edu.au}}
}
%\end{comment}
%******************

\maketitle

\begin{abstract}
Few-shot class-incremental learning (FSCIL) aims to incrementally fine-tune a model (trained on base classes) for a novel set of classes using a few examples without forgetting the previous training. Recent efforts address this problem primarily on 2D images. However, due to the advancement of camera technology, 3D point cloud data has become more available than ever, which warrants considering FSCIL on 3D data. This paper addresses FSCIL in the 3D domain. In addition to well-known issues of catastrophic forgetting of past knowledge and overfitting of few-shot data, 3D FSCIL can bring newer challenges. For example, base classes may contain many synthetic instances in a realistic scenario. In contrast, only a few real-scanned samples (from RGBD sensors) of novel classes are available in incremental steps. Due to the data variation from synthetic to real, FSCIL endures additional challenges, degrading performance in later incremental steps. We attempt to solve this problem using Microshapes (orthogonal basis vectors) by describing any 3D objects using a pre-defined set of rules. It supports incremental training with few-shot examples minimizing synthetic to real data variation. We propose new test protocols for 3D FSCIL using popular synthetic datasets (ModelNet and ShapeNet) and 3D real-scanned datasets (ScanObjectNN and CO3D). By comparing state-of-the-art methods, we establish the effectiveness of our approach in the 3D domain. Code is available at: \textbf{\texttt{\url{https://github.com/townim-faisal/FSCIL-3D}}}.
.
\end{abstract}

\section{Introduction}

Humans have a remarkable ability to gradually expand their knowledge while keeping past insights intact. Similarly, natural learning systems are incremental in nature, where new knowledge is continually learned over time while preserving existing knowledge  \cite{articleccv}. Therefore, it is important to design systems that have the ability to learn incrementally when exposed to novel data. The above task becomes more challenging in realistic scenarios, where only a few samples of novel classes are available for training. This restriction makes incremental learning further difficult due to two reasons:  \textit{a}) overfitting of novel classes with few training samples and \textit{b}) catastrophic forgetting of the previous knowledge. This problem setting is known as Few-shot class incremental learning (FSCIL) in the literature \cite{il2m2018,scail2020,castro2018,lwf2017}. While incremental learning for image data has been studied to a certain extent, its extension to 3D data remains unexplored. To our knowledge, this is the first work that aims to tackle FSCIL on 3D point clouds.

\begin{figure}[t]
\centering

\includegraphics[width=.9\textwidth,trim={0cm 0cm 0cm 0cm},clip]{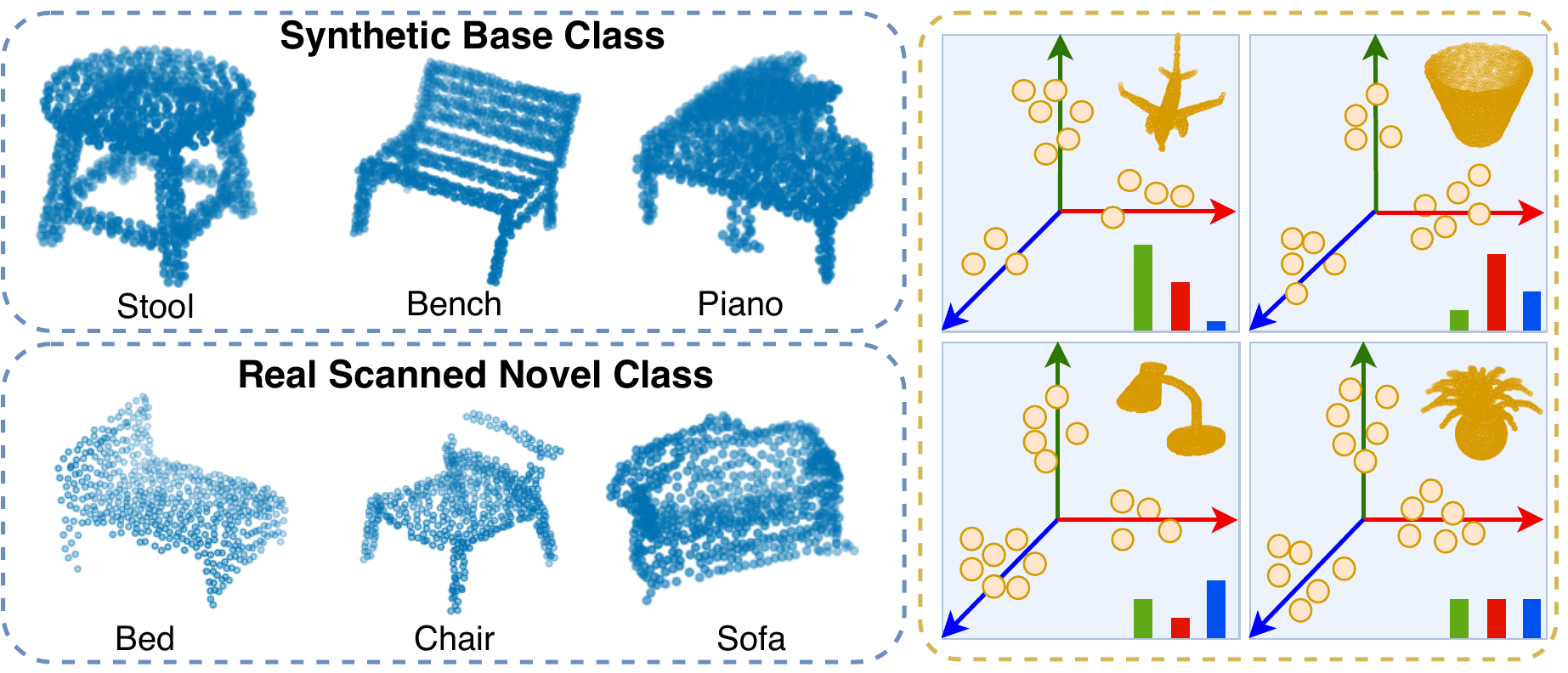}
\caption{\small %Motivational illustration.
\textbf{\textit{(Left)}} A realistic setup of FSCIL may consider synthetic stool, bench, and piano as base classes and real-scanned bed, chair and sofa as novel (few-shot) classes. One can notice that shared `leg' looks (size and shape) different in base and novel classes, which is a source of domain gap. This gap intensified when the synthetic `legs' needed to generalize the real-scanned (noisy) `legs'. Therefore, we propose to describe all `legs' under a common description, named Microshape. Note that Microshapes are abstract property that may not always have semantically conceivable meaning like `legs'. They are orthogonal vectors expressing shared semantics of any object.
\textbf{\textit{(Right)}} High dimensional representations of 3D points are projected onto Microshapes (red, green and blue axis) and calculated to what extent (colored bars) an object contains those Microshapes, which construct a semantic description of the given object.
}
\label{fig:introfig}
\end{figure}
FSCIL methods usually train a base model initially using abundantly available base class instances. Then, new data samples of few-shot incremental tasks are added over time to train the network. During incremental training, models typically tend to overfit few-shot data and forget previously trained class knowledge. Existing works on FSCIL have addressed the latter issue using a small memory module that contains few examples of the previous tasks \cite{tao2020few} (during learning with few-shot examples of novel classes) and the former issue by prototypical descriptions of class representation \cite{Zhu_2021_CVPR}. More recent works~\cite{Cheraghian_2021_ICCV,cheraghian2021semantic} advocated the benefit of using language prototype instead of vision/feature prototype. Inspired by past works, we design our FSCIL method based on a memory module, and link vision and language information in an end-to-end manner. But, FSCIL on 3D entail novel challenges in addition to overfitting and forgetting. Assume a scenario where a robot with 3D sensors  and incremental learning capability is exploring a real environment. The robot is already well-trained with many instances of base classes during its construction. Because of the abundant availability, one can consider using synthetic point cloud instances to train the robot. However, the robot can only obtain a few samples of real-world scanned data for incremental stages. It is a challenge for the robot trained on base classes (with synthetic data) to gather knowledge of incremental classes with  newly scanned real-world data. Previous works \cite{scanobjectnn19} showed that synthetic data-based training and real-world  data-based testing leads to a significant performance gap because of the variability of feature distribution (domain gap) of objects (see  Fig. \ref{fig:introfig}). Similarly, the transition from the synthetic data dependant base task to real-world data dependant incremental tasks makes feature-prototype alignments cumbersome during the incremental steps, eventually magnifying forgetting and overfitting issues.  As a result, the models experience a drastic drop in performance in the  subsequent incremental steps. This paper attempts to address this issue using a set of Microshape descriptions obtained from base classes instances.

Our goal is to represent synthetic and real objects in a way that helps to align 3D point cloud 
features with language prototype of classes. This representation should be robust to noisy real-scanned objects, so that object features describe a meaningful set of attributes. To this end, based on many base class instances, we calculate a set of Microshape descriptions (vectors orthogonal to each other) working as a building block for any 3D shape. Each Microshape describes a particular aspect of a 3D shape that might be shared across different objects. We project a high dimensional representation of 3D points onto each Microshape and find to what extent each Microshape is present in the given object. Aggregating (average pool) the strength of all Microshape presented in a given shape, we create a single feature vector describing that 3D shape. Representing all class objects under a common set of Microshapes, the overall object representation becomes relatively more noise-tolerant, minimizes the domain gap from synthetic to real data, and better aligns features to prototypes. 
Also, such representation can be equally useful for related problem like 3D recognition and dynamic few-shot learning. Moreover, we have proposed new experimental testbeds for testing FSCIL on 3D point cloud objects based on two 3D synthetic datasets, 
ModelNet \cite{modelnet2015} and ShapeNet \cite{shapenet15}, 
and two 3D real world-scanned datasets
, ScanObjectNN \cite{scanobjectnn19} and Common Objects in 3D (CO3D) \cite{co3d21}
. We benchmark popular FSCIL methods on this new setup and show the superiority of our proposed Microshape based method over the existing works. 

In summary, the main contributions of our paper are as follow: \textit{\textbf{(a)}} To the best of our knowledge, we are the first to report few-shot class incremental learning results for 3D point cloud objects. Moreover, we propose a novel and realistic problem setup for 3D objects where base classes are originated from synthetic data, and later few-shot (novel) classes obtained from real-world scanned objects are incrementally added over time (in future tasks). The motivation is that synthetic 3D point cloud objects may not be available for rare class instances, whereas few real-world scanned data may be readily available.
\textit{\textbf{(b)}} We propose a new backbone for feature extraction from 3D point cloud objects that can describe both synthetic and real objects based on a common set of descriptions, called Microshapes. It helps to minimize the domain gap between synthetic and real objects retaining old class knowledge.
\textit{\textbf{(c)}} We propose experimental testbeds for 3D class-incremental learning based on two 3D synthetic datasets, and two 3D real-scanned datasets. We also perform extensive experiments using existing and proposed methods by benchmarking performances on the proposed setups.

\section{Related work}

\noindent\textbf{3D point cloud object recognition:} Thanks to recent advent of 3D sensors \cite{korres2016haptogram,schumann2018semantic,yue2018lidar}, many works have been proposed to classify 3D point cloud objects directly ~\cite{pointnet2017,pointnet++2017,RS-CNN2019,spidercnn2018,SPHNet2019,SFCNN2019,pointconv2019,pointcnn2018}. Qi~\etal{}~\cite{pointnet2017} proposed Pointnet, as the pioneer work, to process 3D point cloud with multi-layer perceptron (MLP) networks. This approach ignores local structures of the input data. Later methods are developed to handle this limitation~\cite{pointnet++2017,RS-CNN2019,spidercnn2018,SPHNet2019,SFCNN2019,pointconv2019,pointcnn2018}. \cite{pointnet++2017} introduced PointNet++ to take advantages of local features by extracting features hierarchically. \cite{RS-CNN2019,spidercnn2018,SPHNet2019,SFCNN2019,pointconv2019,pointcnn2018} proposed several convolution strategies to extract local information. \cite{pointconv2019} introduced PointConv to define convolution operation as locating a Monte Carlo estimation of the hidden continuous 3D convolution with respect to an important sampling strategy. \cite{SFCNN2019} introduced graph convolution on spherical points, which are projected from input point clouds. Some other works \cite{dgcnn2019,LocalSpecGCN2018,PointGCN2018} consider each point cloud as a graph vertex to learn features in spatial or spectral domains. \cite{dgcnn2019} proposed DGCNN to construct a graph in feature space and update it dynamically using MLP for each edge. \cite{PointGCN2018} suggested a method to construct the graph using k-nearest neighbors from a point cloud to capture the local structure and classify the point cloud. As the points in 3D point clouds represent positional characteristics, \cite{zhao2021point} proposed to employ self-attention into a 3D point cloud recognition network. \cite{mazur2021cloud} suggested replacing the explicit self-attention process with a mix of spatial transformer and multi-view convolutional networks. To provide permutation invariance, Guo \etal{} \cite{guo2021pct} presented offset-attention with an implicit Laplace operator and normalizing refinement. In this paper, we propose a novel permutation invariant feature extraction process. %with the help of microshape description.

\noindent\textbf{Incremental learning:} Incremental learning methods are separated into three categories, task-incremental ~\cite{Chaudhry_2018_ECCV,riemer2018learning,v.2018variational}, domain-incremental ~\cite{pmlr-v70-zenke17a,dgr2017}, and class-incremental ~\cite{icarl2017,castro2018,Hou_2019_CVPR,Wu_2019_CVPR} learning. Here, we are interested in the class-incremental learning. Rebuffi~\etal~\cite{icarl2017} used a memory bank called ``episodic memory", of the old classes. For novel classes, the nearest-neighbor classifiers are incrementally accommodated. Castro~\etal~\cite{castro2018} employed a knowledge distillation cost function to store information of the previously seen classes and a classification cost function to learn the novel classes. Hou \etal~\cite{Hou_2019_CVPR} introduced a novel method for learning a unified classifier that reduces the imbalance between old and novel classes. %They take advantage of cosine similarity. 
Wu \etal~\cite{Wu_2019_CVPR} fine-tuned the bias in the model's output with the help of a linear model. They proposed a class-incremental learning method working on the low data regime. Simon \etal{} \cite{Simon_2021_CVPR} presented a knowledge distillation loss using geodesic flow to consider gradual change among incremental tasks. Liu \etal{} \cite{Liu_2021_CVPR} introduced Adaptive Aggregation Networks to aggregate two feature outputs from two residual blocks at each residual level achieving a better stability-plasticity trade-off. Zhu \etal{} \cite{Zhu_2021_CVPR} proposed a non-exemplar-based method consisting of prototype augmentation and self-supervision to avoid memory limitations and task-level overfitting. % This paper applies class-incremental learning on 3D data.

\noindent\textbf{Few-shot class-incremental learning:} FSCIL problem setting was proposed by Tao \etal~\cite{tao2020few}. They proposed a neural gas network to minimize the forgetting issue by learning and preserving the topology of the feature from different classes. Chen \etal~\cite{chen2021incremental} suggested a non-parametric method to squeeze knowledge of the old tasks in a small quantized vector space. They also consider less forgetting regularization, intra-class variation, and calibration of quantized vectors to minimize the catastrophic forgetting. After that, \cite{mazumder2021few} introduced a method that selects a few parameters of the model for learning novel classes to reduce the overfitting issue. Also, by freezing the important parameters in the model, they prevent catastrophic forgetting. \cite{cheraghian2021semantic} employed class semantic information from text space with a distillation technique to mitigate the impact of catastrophic forgetting. Additionally, they utilize an attention mechanism to reduce the overfitting issue on novel tasks, where only a tiny amount of training samples are available for each class. \cite{Cheraghian_2021_ICCV} proposed a method that creates multiple subspaces based on the training data distribution, where each subspace is created based on one specific part of the training distribution, which ends to unique subspaces. The preceding approaches were only explored on 2D image data, whereas our method investigates FSCIL in 3D point cloud data.
\begin{figure} [!t]
\centering
\includegraphics[width=.85\linewidth,trim=.4cm .2cm .3cm .0cm, clip]{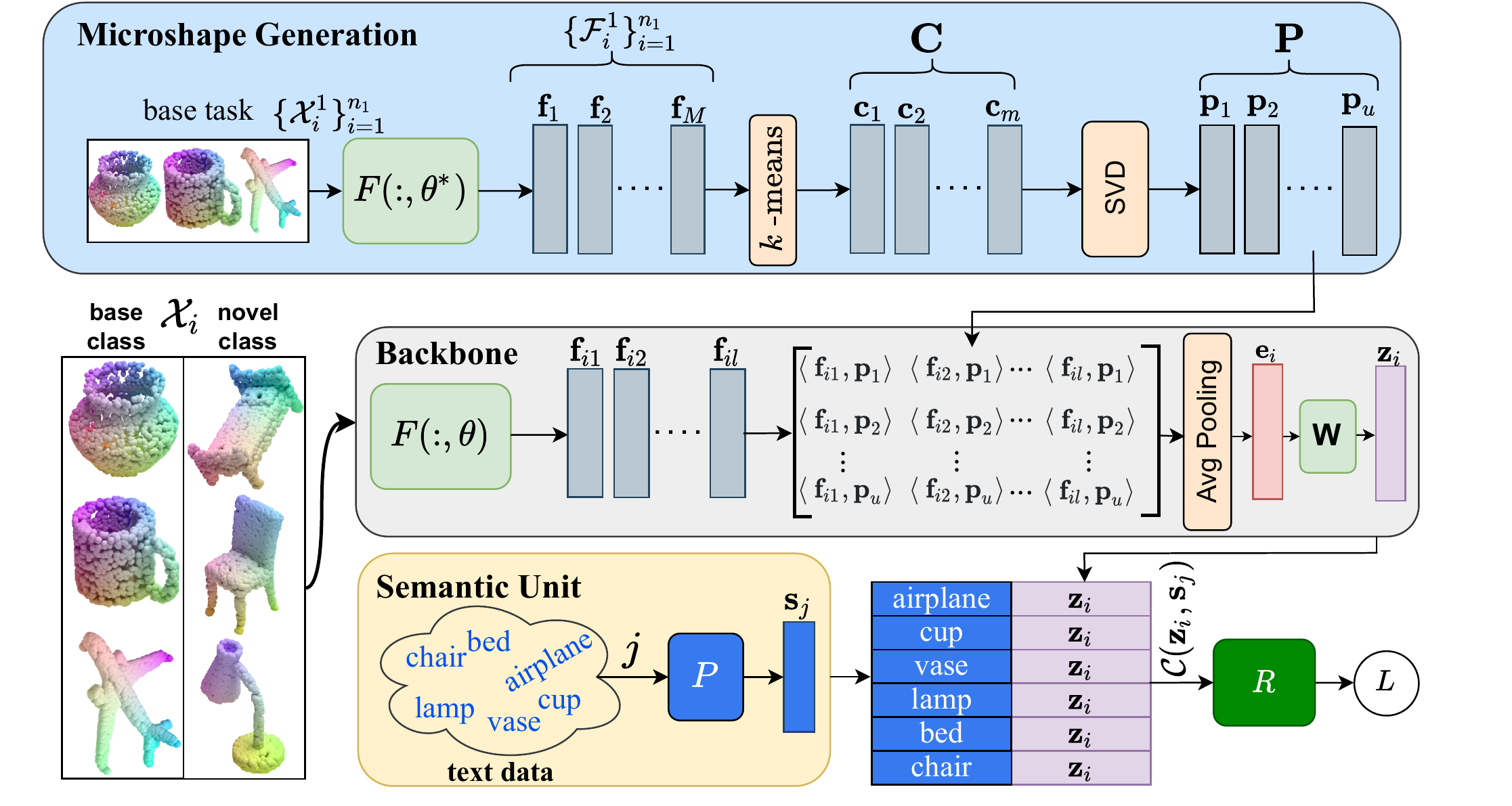}
\caption{\small Overall architecture. \textbf{Microshape Generation:} %In this pipeline, we present the Microshape generation procedure. 
All training samples of the base task are used to calculate Microshapes. To this end, the base training set $\{ \mathcal{X}^1_i \}_{i=1}^{n_1}$ are fed into $F(:,\theta^*)$ in order to extract features for all points $\{ \mathcal{F}^1_i \}_{i=1}^{n_1}$. Next, we apply $K$-means to find $m$ centers $\textbf{C} \in
\mathbb{R}^{q\times m}$ of the point cloud features. Ultimately, we employ SVD to form Microshape matrix $\textbf{P} \in
\mathbb{R}^{q\times u}$. \textbf{Backbone:} Our proposed backbone for generating features of point cloud input is presented. To be more specific, given an instance from 3D point cloud data $\mathcal{X}_{i}$, we map it into a higher dimensional space using $F(:,\theta)$. After that, we calculate the inner product between all points and Microshape basis $\langle\ \textbf{f}_{i},\textbf{p}_{k}\rangle$, where $\textbf{f}_{i}, \textbf{p}_{k} \in \mathbb{R}^{q}$. At the end, we obtain the feature embedding $\textbf{z}_{i} \in
\mathbb{R}^{d}$ after Avg. pooling and a projection $\textbf{W} \in
\mathbb{R}^{d\times u}$ module. $F(:,\theta)$ and $\textbf{W}$ are trainable at the first incremental step but remain fixed in later steps. \textbf{Semantic Unit:} Semantic prototypes, $\textbf{s}_{j}$ of all novel and base classes are generated by a language model, $P$. At the end, a pair of point cloud features and semantic prototype of classes are provided into the relation module, which calculates a class label based on the similarity score. 
}
\label{fig:baseline}
\end{figure}
\section{Few-shot Class-incremental Learning for 3D}
\noindent\textbf{Problem formulation.}
FSCIL models learn novel classes with a small amount of training samples over time in a variety of (incremental) tasks. Each task contains novel classes which have not been observed by the model beforehand. Assume a sequence of $T$ tasks, $\mathcal{H} = \{h^1, h^2, ..., h^T\}$, where $\mathcal{Y}^{t}$ is the set of classes in the task $h^{t}$, and $\mathcal{Y}^{i}\cap \mathcal{Y}^{j}= \emptyset$. In addition, a set of $d$-dimensional semantic prototype for each class of all tasks are assigned as $\mathcal{S}^{t}$. That is, each task can be represented with a tuple $h^t = \left \{\mathcal{X}^{t}_{i},\textbf{y}^{t}_{i},\textbf{s}^{t}_{i} \right \}_{i=1}^{n_{t}},$ where, $\mathcal{X}^{t}_{i} = \{\textbf{x}_{i}^{t}\}_{i = 1}^{l}$ denotes a 3D point cloud object with coordinates  $\textbf{x}_{i}^{t}\in\mathbb{R}^{3}$. Further, $\textbf{y}^{t}_{i} \in \mathcal{Y}^{t}$  and  $\textbf{s}^{t}_{i} \in \mathcal{S}^{t}$ are the label of the point cloud and the corresponding semantic embedding feature, respectively. In the proposed FSCIL setting, $h^1$ is the base task ($t=1$) where the model is trained on a large scale synthetic 3D dataset. For $t>1$, the training data are sampled from real world 3D point clouds, and only consist of few instances, \textit{i.e.}, $n^t \ll n^1$.
The model is trained sequentially over the tasks $t = 1, \dots, T$. However, during $t$-th task, the model sees $\mathcal{X}^t$,  $y^t$ and 
$\{\mathcal{S}^1, \mathcal{S}^2, ..., \mathcal{S}^t\}$.
At inference, the trained model on the current task $h^{t}$ must classify test samples of current and old tasks \textit{i.e.}, $\{h^1, h^2, ..., {h}^t\}$.

\subsection{Model Overview}

To solve the FSCIL problem on the 2D domain, we need to design an algorithm that can address the catastrophic forgetting of old classes and the overfitting issue on novel classes~\cite{tao2020few}. Likewise, we require to follow the same procedure to develop an FSCIL model for 3D point cloud data. In this paper, we use the exemplar approach~\cite{il2m2018,castro2018,scail2020,icarl2017,Hou_2019_CVPR}, which is an established method in the literature, to address the forgetting issue. To this end, for old tasks, we save a sample, chosen randomly for each class in a tiny memory $\mathcal{M}$. Additionally, we employ semantic embedding features in our proposed baseline model to tackle the overfitting issue. Such semantics have been used recently~\cite{Cheraghian_2021_ICCV,cheraghian2021semantic} for the FSCIL in the 2D domain. Furthermore, incremental learning from synthetic (base task) to real-world scanned 3D point cloud data (novel task), enlarges forgetting and overfitting issues. Therefore, the model performance drops drastically in subsequent increment steps. To address this problem, we propose a novel method, Microshape, which is shown in Fig.~\ref{fig:baseline}. In this method, all training samples of the base task  $\{ \mathcal{X}^1_i \}_{i=1}^{n_1}$ are projected into a higher dimension space by $F(:,\theta^*)$, which is a pretained model on the base task by a cross-entropy loss and consist of a few MLP layers, leading to $\{ \mathcal{F}^1_i \}_{i=1}^{n_1}$. Next, we calculate $m$ class centers $\textbf{C} \in \mathbb{R}^{q \times m}$ by K-means algorithm. At the end, Singular Value Decomposition (SVD) is used to choose the most important Microshapes~$\textbf{P} \in
\mathbb{R}^{q\times u}$, where $u < m$. The details of the Microshape generation algorithm are explained in Sec.~\ref{sec_Microshape}. After this step, we train our proposed architecture for 3D point cloud data, shown in Fig.~\ref{fig:baseline}, with the help of Microshapes for base and novel tasks. In the point cloud pipeline, there are two kinds of classes: novel and old. It is important to mention that the old class samples are from the tiny memory $\mathcal{M}$. In the text pipeline, there is one representation per class for old and novel classes. While training the baseline model, we forward a 3D point cloud sample, $\mathcal{X}_{i}$, into the projection module $F(:,\theta)$ to extract point cloud features $\mathcal{F}_{i}=\{ \textbf{f}_i \}_{i=1}^{n}$, where $\textbf{f}_{i} \in \mathbb{R}^{q}$. Then, we find the inner product $\langle\,,\rangle$ between all point cloud features $\textbf{f}_{i}$ and Microshape $\textbf{p}_{k}\in \mathbb{R}^{q}$. After that, we use average pooling for each Microshape to form a feature vector representation $\textbf{e}_{i} \in \mathbb{R}^{u}$ of the point cloud input $\mathcal{X}_{i}$. Next, we map the feature vector $\textbf{e}_{i}$ into semantic embedding space by $\textbf{W} \in  \mathbb{R}^{d \times u}$, which is a fully connected layer, to form $\textbf{z}_{i} \in \mathbb{R}^{d}$  . In the text domain, all current and previous class labels are fed into a semantic embedding module $P$, which can generate a feature embedding $\textbf{s}_{j}\in \mathbb{R}^{d}$ for each class. Ultimately, we forward $\textbf{z}_{i}$ and $\textbf{s}_{j}$ in the relation network~\cite{8578229} $R$, consist of a few fully connected layers, which gives a prediction given input $\mathcal{X}_{i}$. In the end, we use a binary cross-entropy loss to train the model. It is important to mention that $F(:,\theta)$ and $\textbf{W}$ are trained only on the base task and kept frozen for the incremental tasks. While the relation module $R$ is fine-tuned for all incremental tasks.

\subsection{Microshapes} 
\label{sec_Microshape}
Incremental learning faces newer challenges, from 2D images to 3D point cloud objects. \textit{First,} it is more difficult to obtain noise-free real-world 3D data reconstructed from RGB-D scan than natural 2D images. 3D objects are susceptible to partial observations, occlusions, reconstruction errors, and cluttered backgrounds. 
An effect of the noise-free version of 3D (synthetic) and real-world scanned objects is shown in Fig \ref{fig:introfig} (left).
\textit{Second,} it is realistic to assume that 3D (synthetic) objects are relatively abundant for base classes but scarce for novel classes. A practical incremental learning setup should consider synthetic objects for training base classes (1st incremental step) and real-scanned objects for the rest of the incremental steps containing few-shot classes. The 3D-feature and language-prototype alignment learned using base instances could not generalize well for real-scanned data of novel classes. It increases the domain gap between base and novel classes. Now, we formally present the following hypothesis.

\noindent\textbf{Hypothesis 1.} \textit{Every 3D shape can be adequately represented  as a particular combination of smaller entities. These entities are common across various 3D shapes, while the combinations might vary.}

\noindent Based on the above hypothesis, we strive to mitigate the domain gap between synthetic and real 3D point cloud data by proposing a novel feature extraction method, called Microshapes. In the proposed method, the common properties of samples are selected to form a feature representation which is robust to domain shift between synthetic and real data. The assumption is that there are many similar points among different point cloud objects that can form as a cluster and can be termed as Microshape. After that, we extract feature representation of all samples based on Microshapes. 

\noindent\textbf{Microshape generation.} Let $\{ \mathcal{X}^1_i \}_{i=1}^{n_1}$ be the set of 3D samples used to train the model in the base task $h^1$. 
We forward $\{ \mathcal{X}^1_i \}_{i=1}^{n_t}$ through the backbone $F$ and extract $M$ features $\{\textbf{f}_{i}\}_{i=1}^M$, where $f_i \in \mathbb{R}^q$, $M = l \times n_1$, and $l$ is the number of points in a point cloud. For the backbone, we use a pre-trained PointNet (trained using $\{ \mathcal{X}^1_i \}_{i=1}^{n_t}$), after removing the pooling and classification layers. Then, we obtain $m$ cluster centers by applying K-means on the set $\{\textbf{f}_{i}\}_{i=1}^M$. Let us denote the cluster centers by $\textbf{C} \in \mathbb{R}^{q \times m}$. Our intention is to define a vector space spanned by the column vectors of $\textbf{C}$ which then can be used to represent novel 3D point clouds as a vector in this space. However, the column vectors of $\textbf{C}$ is not necessarily a basis since they might not be linearly independent. On the other hand, the number of cluster centers $m$ is a hand picked hyperparameter, which might not be optimal. Therefore,  we use Singular  Value Decomposition (SVD) to choose a set of basis vectors that approximately spans $\mathbf{C}$.
We decompose the matrix consisting of samples within a cluster as $\textbf{C} = \textbf{U}\textbf{D}\textbf{V}^{\top}$. Then, the $u$ leading left singular vectors $\textbf{U}$
form an orthogonal basis which we define by $\textbf{P}$, \textit{i.e},
$\mathbb{R}^{q \times u} \ni \textbf{P} = \left [ \textbf{p}_{1},...,\textbf{p}_{u} \right ];~\textbf{P}^{\top}\textbf{P}=\mathbf{I}_{n}$ (see Fig.~\ref{fig:baseline}).

% \begin{definition}
\noindent\textbf{Definition 1.} \textit{We define each column vector of $\textbf{P}$ as a Microshape.} 
% \end{definition}

\noindent \textbf{Role of SVD.} Since the Microshapes are mutually orthogonal, they contain minimal mutual information, which allows us to encode maximum amount of total information within them. In comparison, the raw cluster centers obtained using k-means might be linearly dependant, representing duplicate information. Choosing orthogonal basis vectors based on the 95\% of energy of singular values allows to find the number of Microshapes required to describe a dataset in a principled manner. Further, since the number of cluster centers is a hand picked parameter, they can result in either redundant or insufficient information encoding. In contrast, using SVD allows us eliminate the effect of this hyperparameter.

\noindent\textbf{Microshape feature extraction.} 
\label{sec:microshape_feature}
To form a feature vector that describes a particular 3D shape, we use these Microshapes. Consider a point cloud $\mathcal{X}_{i}=\{\textbf{x}_{i1},...,\textbf{x}_{il}
\}$. We project it into a higher dimensional space using $F$, which gives  $\mathcal{F}_{i}=\{\textbf{f}_{i1},...,\textbf{f}_{il}
\}$.  Next, for each Microshape, we calculate the average similarity value to create the feature vector,
\begin{align}
    \mathbf{e}_{i} = \frac{1}{l}\sum_{b=1}^l\left[ \textbf{f}_{ib} \cdot \textbf{p}_1, \textbf{f}_{ib} \cdot \textbf{p}_2, \dots, \textbf{f}_{ib} \cdot \textbf{p}_u \right], \;\;\;\;\;\;\; \mathbf{z}_{i}  = ReLU(\textbf{W} . \mathbf{e}_{i}),
    \label{eq:avgpool}
\end{align}
where, $\mathbf{e}_{i} \in \mathbb{R}^{u}$. Note that $\mathbf{e}_{i}$ is a permutation invariant representation, since it is calculated using all the projected points $\textbf{f}_i$'s of the original point cloud, and is independent of the point ordering. An FC layer ($\textbf{W}$) converts $\mathbf{e}_{i}$ to $\mathbf{z}_{i} \in \mathbb{R}^{d}$. Next, we describe our training pipeline utilizing the obtained $\mathbf{z}_{i}$ of each objects.

\noindent\textbf{Benefit of Mircoshape features.} FSCIL allows only a few examples of real-scanned objected during incremental stages. Inherent noise/occlusion/clutter background presented in 3D models results in overfitting to unnecessary information the incremental learning steps. In this context, Microshpae based feature representation has benefits over general features such as PointNet representations. Below, we list several key benefits of  Mircoshape features: 
\textit{\textbf{(a)} Reduced impact of noise:} Microshapes allows the model to describe any object based on common entities. As explained earlier, traditional backbone (such as Pointnet) features can represent redundant information. In contrast, Microshapes only try to estimate to what extent these pre-defined properties are present in a given input and  helps to discard unnecessary information. Such an approach is beneficial to describe novel objects, especially in cross-domain (synthetic to real) situations.
\textit{\textbf{(b)} Aligning with the prototypes:} We use language prototypes that contain semantic regularities such that similar concepts like (table, chair, etc.) are located in nearby positions in a semantic space. By describing objects based on their similarity to pre-defined Microshapes, features can describe different semantics/attributes of a given object. As language prototypes and 3D features are based on object semantics rather than low-level (CNN) features, aligning them becomes an easier task for the relation network.
\textit{\textbf{(c)} Microshapes as feature prototypes:} For any few-shot learning problem, Microshapes can serve the role of creating feature prototypes (instead of language prototypes used in this paper). One can consider the average of Microshape-based features of all available instances belonging to any class, as a class prototype. This could be useful to describe fine-grained objects (e.g., China/Asian Dragon, Armadillo) for which language prototypes are less reliable. In the experiments, we deal with common objects where language prototypes work better than feature prototypes. 
% (see Table \ref{table:ablation_study_cluster_w2v}).

\subsection{Training pipeline}

First, we train the backbone $F$ on the base task $h^1$, with a large number of synthetic 3D samples. For the novel tasks, $h^{t > 1}$, the backbone $F$ is kept frozen. The semantic class backbone $P$, which is a pretrained model, \textit{e.g} BERT~\cite{devlin-etal-2019-bert} and w2v~\cite{mikolov2013distributed}, is kept frozen during the entire training stage. Additionally, for each old class, we randomly select a training sample which is stored on a tiny memory $\mathcal{M}$. Let us define the feature vector obtained using Eq.~\ref{eq:avgpool} for the $i^{th}$ point cloud of the $t^{th}$ task as $\textbf{z}_{i}^{t}$. To train the proposed model for a task $h^t$, at first, we generate the features $\textbf{z}_{i}^{t}$'s using Eq.~\ref{eq:avgpool} for the $n_t$ training samples of the current task $h^{t}$. Then, the features $\textbf{z}_{i}^{t}$ and semantic class embedding $\textbf{s}_{j}$ of the tasks $h^1, \dots, h^t$ are forwarded into a relation module $R$ which provides a score between $\left[ 0,1 \right]$, representing the similarity between them. In other words, for each training sample, we generate a score against each of the classes in both the novel and previous tasks as, $r_{ij}^t = \gamma \circ R \circ (\textbf{z}^t_{i} \oplus {\textbf{s}}_{j}) , j \in \mathcal{Y}_{tl}$
where, $\mathcal{Y}_{tl} = \bigcup_{i=1}^{t} \mathcal{Y}^{i}$,  $\oplus$ is the concatenation operator, $R$ is the relation module, and $\gamma$ is the sigmoid function. 
Finally, for each feature $r_{ij}$, and the corresponding ground truth $\textbf{y}_i$, we employ a binary cross entropy cost function to train the model as,
\begin{align}
    L=-\frac{1}{|\mathcal{Y}_{tl}||\mathcal{S}|}\sum_{j \in \mathcal{Y}_{tl}}^{}\sum_{y_{i}\in\mathcal{S}}\Bigg( \mathbb{1}_{(y_{i} = j)}\mathrm{log}(r_{ij}) + (1 - \mathbb{1}_{(y_{i} = j)})\mathrm{log}(1 - r_{ij}) \Bigg)
     \label{eqn:total_loss_1}
\end{align}
where $\mathcal{S}$ is the set of true labels in the current task and the memory $\mathcal{M}$.

\noindent \textbf{Inference.} During inference, given the trained model an unlabeled sample $\mathcal{X}^{c}$, $c \in \mathcal{Y}_{tl}$, the prediction of the label is calculated by $y^{*} =\underset{j \in \mathcal{Y}_{tl}}{\arg \max} \, \, \big( R \circ  (\textbf{z}_c \oplus \textbf{s}_{j}) \big),$
% \begin{align}
% % $$
% y^{*} =\underset{j \in \mathcal{Y}_{tl}}{\arg \max} \, \, \big( R \circ  (\textbf{z}_c \oplus \textbf{s}_{j}) \big)\;,
% % $$
% \label{eqn:inference}
% \end{align}
where $\textbf{z}_c$ is the feature calculated using the microshapes for the sample $\mathcal{X}^c$.

\begin{table}[!t]
\newcolumntype{C}[1]{>{\centering\arraybackslash}p{#1}}
\centering \small
\caption{\small Summary of our experimental setups. 
}
\scalebox{0.7}{
% \begin{minipage}{0.8\linewidth}
% \centering \small
\begin{tabular} 
{C{5cm}|C{1.4cm}C{1.4cm}C{1.4cm}C{1.4cm}C{1.4cm}C{1.4cm}}
% {C{5cm}|C{1.8cm}C{1.8cm}C{1.8cm}C{1.8cm}C{1.8cm}C{2.2cm}}
\hline
\multirow{2}{*}{Experiment Setups} & \# Base  & \# Novel  & \multirow{2}{*}{\# Tasks} & \#Train  & \#Test  & \#Test \\
 & Classes  & Classes &  & in Base &  in Base &  in Novel\\
\hline
ModelNet40 & 20 & 20 & 5 & 7438 & 1958 & 510\\
ShapeNet & 25 & 30 & 7 & 36791 & 9356 & 893\\
CO3D & 25 & 25 & 6 & 12493 & 1325 & 407\\
ModelNet40 $\rightarrow$ ScanObjectNN & 26 & 11 & 4 & 4999 & 1496 & 475\\
ShapeNet $\rightarrow$ ScanObjectNN & 44 & 15 & 4 & 22797 & 5845 & 581\\
ShapeNet $\rightarrow$ CO3D & 39 & 50 & 11 & 26287 & 6604 & 1732\\
\hline
\end{tabular}
% \end{minipage} 
}
\label{table:experimental_setup}
\end{table}

\section{Experiments}

\textbf{Datasets.} We experiment on two 3D synthetic datasets, (ModelNet \cite{modelnet2015}, ShapeNet \cite{shapenet15}) and two 3D real-scanned datasets (ScanObjectNN \cite{scanobjectnn19} and CO3D \cite{co3d21}).

\noindent\textbf{Setups.} We propose two categories of experiments for 3D incremental learning: within- and cross-dataset experiments. Within dataset experiments are done on both synthetic and real datasets individually where the base and incremental classes come from the same dataset. For cross-dataset experiments, base and incremental classes come from synthetic and real-world scanned datasets, respectively. In Table \ref{table:experimental_setup}, we summarize three within and three cross-dataset experimental setups proposed in this paper. 
We utilize the class distribution of ModelNet, ShapeNet, and CO3D datasets to select base and incremental few-shot classes for within dataset experiments. First, we sort all classes in descending order based on instance frequency. 
Then top 50\% of total classes (having many available instances) are chosen as base classes, and the rest (having relatively fewer available examples) are added incrementally in our experiments. The motivation is that rare categories are the realistic candidate for novel (few-shot) classes. In the ModelNet40 experiment, there is one base task with 20 classes and four incremental tasks with another 20 classes. ShapeNet and CO3D experiments have 25 base classes, whereas incremental classes are 30 and 25, respectively, divided into 6 and 4 tasks.
Among cross-dataset experimental setups, we choose base classes from a synthetic dataset and later add incremental classes from a real-world scanned dataset. 
For ModelNet40 $\rightarrow$ ScanObjectNN experiment, we follow the selection of base and incremental (novel) classes  from \cite{cheraghian2021zero}, and it has total 4 tasks. ShapeNet $\rightarrow$ ScanObjectNN has four tasks where 44 non-overlapping ShapeNet and 15 ScanObjectNN classes are used as the base and incremental classes, respectively. ShapeNet $\rightarrow$ CO3D experiment has a sequence of 11 tasks with 44 non-overlapped classes from ShapeNet as base classes and 50 classes from CO3D as incremental classes. This setup is the most challenging and realistic among all experiments because of its vast number of tasks, classes, and object instances. Unless mentioned explicitly, we use randomly selected five 3D models as few-shot samples/class in all FSCIL setups and allow one exemplar 3D model/class (randomly chosen) as the memory of previous tasks. For synthetic and real-scanned object classes, we use synthetic and real-scanned data as few-shot or exemplar data, respectively. See the supplementary material for more details about the setups.\\
\noindent\textbf{Semantic embedding.} We use 300-dim. word2vec \cite{mikolov2013distributed} as semantic representation for each classes of datasets. The word2vec is produced from word vectors trained in an unsupervised manner on an unannotated text corpus, Wikipidia.\\
\noindent\textbf{Validation strategy.} We make a validation strategy for within dataset experiments. We randomly divide the set of base classes into val-base and val-novel. We choose 60\% classes from base classes as val-base classes and the rest as val-novel classes to find the number of centroids for generating Microshapes. We find the number of centroids, $m=1024$ performing across experimental setup.\\
\noindent\textbf{Implementation details.} In all experiments and compared approaches, we use PointNet \cite{pointnet2017} as a base feature extractor to build the centroids for microshapes. We use the farthest 1024 points from 3D point cloud objects as input for all samples. We train the base feature extractor for 100 epochs using Adam optimizer with a learning rate of 0.0001. For microshape formulation, we use $k$-means clustering algorithms to initialize centroids. We randomly shift and scale points in the input point cloud during base and incremental class training with randomly dropping out points. During the training of all base classes, we employ the Adam optimizer with a learning rate of 0.0001 and batch sizes of 64. For novel classes, we choose the batch size of 16 and the learning rate of 0.00005. The feature vector size from backbone is 300 dimensional, similar to the dimension of the semantic prototypes. In the relation network, we utilize three fully connected layers of (600,300,1) using LeakyReLU activations, except the output layer uses Sigmoid activation. We use the \textit{PyTorch} framework to perform our experiments.\\
\noindent\textbf{Evaluation metrics.} We calculate the accuracy after each incremental step by combining both base and novel classes. Finally, as suggested in \cite{graph-few-shot2022}, we calculate the relative accuracy dropping rate, $\Delta  = \frac{|acc_{T}-acc_{0}|}{acc_{0}} \times 100$, where, $acc_{T}$ and $acc_{0}$ represent the last and first incremental task's accuracy, respectively. $\Delta$ summarizes the overall evaluation of methods. Lower relative accuracy indicates a better performance. We report the mean accuracy after ten different runs with random initialization.

%%%%% WITHIN DATASET %%%%%
\begin{table}[t]
\newcolumntype{C}[1]{>{\centering\arraybackslash}p{#1}}
\centering \small
\caption{\small Overall FSCIL results for within dataset experiments}
\scalebox{0.65}{
\begin{minipage}{.58\textwidth}
% \caption*{ModelNet}
\begin{tabular}{C{2cm}|C{0.7cm}C{0.7cm}C{0.7cm}C{0.7cm}C{0.7cm}C{0.7cm}}
\hline
& \multicolumn{6}{c}{ModelNet}\\
\hline
Method & 20 & 25 & 30 & 35 & 40 & $\Delta\downarrow$\\
\hline
\textit{FT} & 89.8 & 9.7 & 4.3 & 3.3 & 3.0 & 96.7\\
\textit{Joint} & 89.8 & 88.2 & 87.0 & 83.5 & 80.5 & 10.4\\
\hline
LwF \cite{lwf2017} & 89.8 & 36.0 & 9.1 & 3.6 & 3.1 & 96.0\\
IL2M \cite{il2m2018} & 89.8 & 65.5 & 58.4 & 52.3 & 53.6 & 40.3 \\
ScaIL \cite{scail2020} & 89.8 & 66.8 & 64.5 & 58.7 & 56.5 & 37.1\\
EEIL \cite{castro2018} & 89.8 & 75.4 & 67.2 & 60.1 & 55.6 & 38.1\\
FACT \cite{zhou2022forward} & 90.4 & 81.3 & 77.1 & 73.5 & 65.0 & 28.1\\
Sem-aware \cite{cheraghian2021semantic} & 91.3 & 82.2 & 74.3 & 70.0 & 64.7 & 29.1\\

Ours & \textbf{93.6} & \textbf{83.1} & \textbf{78.2} & \textbf{75.8} & \textbf{67.1} & \textbf{28.3}\\
\hline
\end{tabular}
\end{minipage} \hfill 
\begin{minipage}{.48\textwidth}
% \caption*{CO3D}
\begin{tabular}{C{0.7cm}C{0.7cm}C{0.7cm}C{0.7cm}C{0.7cm}C{0.7cm}C{0.7cm}}
\hline
\multicolumn{7}{c}{CO3D}\\\hline
25 & 30 & 35 & 40 & 45 & 50 & $\Delta\downarrow$\\
\hline
76.7 & 11.2 & 3.6 & 3.2 & 1.8 & 0.8 & 99.0\\
76.7 & 69.4 & 64.8 & 62.7 & 60.7 & 59.8 & 22.0\\
\hline
76.7 & 14.7 & 4.7 & 3.5 & 2.3 & 1.0 & 98.7\\
76.7 & 31.5 & 27.7 & 18.1 & 27.1 & 21.9 & 71.4\\
76.7 & 39.5 & 34.1 & 24.1 & 30.1 & 27.5 & 64.1\\
76.7 & 61.4 & 52.4 & 42.8 & 39.5 & 32.8 & 57.2\\
77.9 & 67.1 & 59.7 & 54.8 & 50.2 & 46.7 & 40.0\\
76.8 & 66.9 & 59.2 & 53.6 & 49.1 & 42.9 & 44.1\\
\textbf{78.5} & \textbf{67.3} & \textbf{60.1} & \textbf{56.1} & \textbf{51.4} & \textbf{47.2} & \textbf{39.9}\\
\hline
\end{tabular}
\end{minipage} \hfill 
\begin{minipage}{.5\textwidth}
% \caption*{ShapeNet}
\begin{tabular}{C{0.6cm}C{0.6cm}C{0.6cm}C{0.6cm}C{0.6cm}C{0.6cm}C{0.6cm}C{0.6cm}}
\hline
\multicolumn{8}{c}{ShapeNet}\\\hline
25 & 30 & 35 & 40 & 45 & 50 & 55 & $\Delta\downarrow$\\
\hline
87.0 & 25.7 & 6.8 & 1.3 & 0.9 & 0.6 & 0.4 & 99.5\\
87.0 & 85.2 & 84.3 & 83.0 & 82.5 & 82.2 & 81.3 & 6.6\\
\hline
87.0 & 60.8 & 33.5 & 15.9 & 3.8 & 3.1 & 1.8 & 97.9\\
87.0 & 58.6 & 45.7 & 40.7 & 50.1 & 49.4 & 49.3 & 43.3\\
87.0 & 56.6 & 51.8 & 44.3 & 50.3 & 46.3 & 45.4 & 47.8\\
87.0 & 77.7 & 73.2 & 69.3 & 66.4 & 65.9 & 65.8 & 22.4\\
87.5 & 75.3 & 71.4 & 69.9 & 67.5 & 65.7 & 62.5 & 28.6 \\
87.2 & 74.9 & 68.1 & 69.0 & 68.1 & 66.9 & 63.8 & 26.8\\
\textbf{87.6} & \textbf{83.2} & \textbf{81.5} & \textbf{79.0} & \textbf{76.8} & \textbf{73.5} & \textbf{72.6} & \textbf{17.1}\\
\hline
\end{tabular}
\end{minipage}
}
\label{table:overall_experiment_synth}
\end{table}

%%%%% CROSS DATASET %%%%%
\begin{table}[!t]
\newcolumntype{C}[1]{>{\centering\arraybackslash}p{#1}}
\centering \small
\caption{\small Overall results for cross dataset experiments}
\scalebox{0.65}{
\begin{minipage}{.88\textwidth}
\begin{tabular}{C{2cm}|C{0.6cm}C{0.6cm}C{0.6cm}C{0.6cm}C{0.6cm}C{0.6cm}C{0.6cm}C{0.6cm}C{0.6cm}C{0.6cm}C{0.6cm}C{0.6cm}}
\hline
& \multicolumn{12}{c}{ShapeNet $\rightarrow$ CO3D}\\
\hline
Method & 39 & 44 & 49 & 54 & 59 & 64 & 69 & 74 & 79 & 84 & 89 & $\Delta\downarrow$\\
\hline
\textit{FT} & 81.0 & 20.2 & 2.3 & 1.7 & 0.8 & 1.0 & 1.0 & 1.3 & 0.9 & 0.5 & 1.6 & 98.0\\
\textit{Joint} & 81.0 & 79.5 & 78.3 & 75.2 & 75.1 & 74.8 & 72.3 & 71.3 & 70.0 & 68.8 & 67.3 & 16.9\\
\hline
LwF \cite{lwf2017} & 81.0 & 57.4 & 19.3 & 2.3 & 1.0 & 0.9 & 0.8 & 1.3 & 1.1 & 0.8 & 1.9 & 97.7\\
IL2M \cite{il2m2018} & 81.0 & 45.6 & 36.8 & 35.1 & 31.8 & 33.3 & 34.0 & 31.5 & 30.6 & 32.3 & 30.0 & 63.0\\
ScaIL \cite{scail2020} & 81.0 & 50.1 & 45.7 & 39.1 & 39.0 & 37.9 & 38.0 & 36.0 & 33.7 & 33.0 & 35.2 & 56.5\\
EEIL \cite{castro2018} & 81.0 & 75.2 & 69.3 & 63.2 & 60.5 & 57.9 & 53.0 & 51.9 & 51.3 & 47.8 & 47.6 & 41.2\\
FACT \cite{zhou2022forward} & 81.4 & 76.0 & 70.3 & 68.1 & 65.8 & 63.5 & 63.0 & 60.1 & 58.2 & 57.5 & 55.9 & 31.3\\
Sem-aware \cite{cheraghian2021semantic} & 80.6 & 69.5 & 66.5 & 62.9 & 63.2 & 63.0 & 61.2 & 58.3 & 58.1 & 57.2 & 55.2 & 31.6\\
Ours & \textbf{82.6} & \textbf{77.9} & \textbf{73.9} & \textbf{72.7} & \textbf{67.7} & \textbf{66.2} & \textbf{65.4} & \textbf{63.4} & \textbf{60.6} & \textbf{58.1} & \textbf{57.1} & \textbf{30.9}\\
\hline
\end{tabular}
\end{minipage} \hfill 
\begin{minipage}{.34\textwidth}
\centering \small
\begin{tabular}{C{0.65cm}C{0.65cm}C{0.65cm}C{0.65cm}C{0.65cm}}
\hline
\multicolumn{5}{c}{ ModelNet $\rightarrow$ ScanObjectNN }\\\hline
26 & 30 & 34 & 37 & $\Delta\downarrow$\\
\hline
88.4 & 6.4 & 6.0 & 1.9 & 97.9\\
88.4 & 79.7 & 74.0 & 71.2 & 19.5\\
\hline
88.4 & 35.8 & 5.8 & 2.5 & 97.2\\
88.4 & 58.2 & 52.9 & 52.0 & 41.2\\
88.4 & 56.5 & 55.9 & 52.9 & 40.2\\
88.4 & 70.2 & 61.0 & 56.8 & 35.7\\
89.1 & 72.5 & 68.3 & 63.5 & 28.7\\
88.5 & \textbf{73.9} & 67.7 & 64.2 & 27.5\\
\textbf{89.3} & 73.2 & \textbf{68.4} & \textbf{65.1} & \textbf{27.1}\\
\hline
\end{tabular}
\end{minipage} \hfill 
\begin{minipage}{.3\textwidth}
\begin{tabular}{C{0.65cm}C{0.65cm}C{0.65cm}C{0.65cm}C{0.65cm}}
\hline
\multicolumn{5}{c}{ShapeNet $\rightarrow$ ScanObjectNN}\\\hline
44 & 49 & 54 & 59  & $\Delta\downarrow$\\
\hline
81.4 & 38.7 & 4.0 & 0.9 & 98.9\\
81.4 & 82.5 & 79.8 & 78.7 & 3.3\\
\hline
81.4 & 47.9 & 14.0 & 5.9 & 92.8\\ 
81.4 & 53.2 & 43.9 & 45.8 & 43.7 \\
81.4 & 49.0 & 46.7 & 40.0 & 50.9 \\
81.4 & 74.5 & 69.8 & 63.4 & 22.1\\
82.3 & 74.6 & 69.9 & 66.8 & 18.8\\
81.3 & 70.6 & 65.2 & 62.9 & 22.6\\
\textbf{82.5} & \textbf{74.8} & \textbf{71.2} & \textbf{67.1} & \textbf{18.7}\\
\hline
\end{tabular}
\end{minipage}
} 
\label{table:overall_experiment_real_synth}
\end{table}

%%%%% ABLATION STUDY USING MICOSHAPES, W2V & L2 %%%%%
\begin{table}[t]
\newcolumntype{C}[1]{>{\centering\arraybackslash}p{#1}}
\centering \small
\caption{\small Ablation study on using Microshapes and semantic prototypes.}
\scalebox{0.6}{
\begin{minipage}{.98\textwidth}
\begin{tabular}{C{1.5cm}C{1.5cm}|C{0.61cm}C{0.61cm}C{0.61cm}C{0.61cm}C{0.61cm}C{0.61cm}C{0.61cm}C{0.61cm}C{0.61cm}C{0.61cm}C{0.61cm}C{0.6cm}}
\hline
& & & \multicolumn{11}{c}{ShapeNet $\rightarrow$ CO3D}\\
\hline
Microshape & Prototype & 39 & 44 & 49 & 54 & 59 & 64 & 69 & 74 & 79 & 84 & 89 & $\Delta\downarrow$\\
\hline
No & Feature & 79.4 & 70.5 & 68.1 & 65.5 & 62.9 & 60.7 & 59.1 & 57.6 & 56.2 & 54.3 & 51.8 & 34.8\\
Yes & Feature & 80.4 & 71.9 & 70.2 & 66.1 & 64.6 & 62.5 & 60.6 & 58.4 & 57.7 & 55.0 & 53.4 & 33.6\\
No & Language & 80.6 & 75.9 & 66.3 & 66.1 & 66.0 & 63.9 & 62.8 & 60.0 & 56.5 & 54.1 & 53.6 & 33.5\\
Yes & Language & 82.6 & 77.9 & 73.9 & 72.7 & 67.7 & 66.2 & 65.4 & 63.4 & 60.6 & 58.1 & 57.1 & 30.9\\
\hline
\end{tabular}
\end{minipage} \hfill 
\begin{minipage}{.5\textwidth}
\begin{tabular}{C{0.62cm}C{0.62cm}C{0.62cm}C{0.62cm}C{0.62cm}C{0.62cm}C{0.62cm}C{0.6cm}}
\hline
\multicolumn{8}{c}{ShapeNet}\\\hline
25 & 30 & 35 & 40 & 45 & 50 & 55 & $\Delta\downarrow$\\
\hline
87.4 & 79.1 & 76.9 & 72.9 & 59.9 & 67.1 & 57.0 & 34.8\\
87.5 & 75.6 & 76.3 & 66 & 66.6 & 63.5 & 62.0 & 29.1\\
87.3 & 79.4 & 76.8 & 72.8 & 68.6 & 66.1 & 63.7 & 27.0\\
87.6 & 83.2 & 81.5 & 79.0 & 76.8 & 73.5 & 72.6 & 17.1\\
\hline
\end{tabular}
\end{minipage}
} 
% \vspace{-1.5em}
\label{table:ablation_study_cluster_w2v}
\end{table}

\subsection{Main results}
\textbf{Compared methods.} We compare our method with the following approaches. (1) \textit{Fine-tuning (FT):} It is vanilla fine-tuning wherein each incremental task, the model is initialized with the previous task's weight and only uses a few samples from incremental classes. Note no exemplar is used in this method. 
(2) \textit{Joint:} All incremental classes are jointly trained using all samples belonging to those classes. \textit{FT} and \textit{Joint} are considered as lower and upper bound results, respectively. (3) State-of-the-art methods, e.g., IL2M \cite{il2m2018}, ScaIL \cite{scail2020}, EEIL \cite{castro2018}, LwF \cite{lwf2017}, FACT \cite{zhou2022forward}, and Sem-aware \cite{cheraghian2021semantic}. These methods originally reported results on 2D datasets. We replace CNN features with PointNet features and use their official implementation to produce results for 3D.

\noindent\textbf{Analysis.} We present the comparative results for within and cross dataset experiments in Table \ref{table:overall_experiment_synth} and \ref{table:overall_experiment_real_synth}, respectively. Our observations are as follows: 
\textbf{\textit{(1)}} All methods perform poorly in general in cross dataset experiments in comparison to within dataset case. This is due to the noisy data presented in 3D real datasets.
\textbf{\textit{(2)}} \textit{FT} gets the lowest performance among all experimental setups because of catastrophic forgetting that occurred during training due to not utilizing any memory sample. In contrast, \textit{Joint} achieves the best results because it uses all samples and trains all few-shot classes jointly at a time. This upper bound setup is not FSCIL.
\textbf{\textit{(3)}} Other state-of-the-art approaches (IL2M, ScaIL, EEIL, LwF, FACT, and Sem-aware) could not perform well in most of the experimental setups. IL2M \cite{il2m2018} and ScaIL \cite{scail2020} propose a special training mechanism for 2D image examples. LwF \cite{lwf2017} and EEIL \cite{castro2018} both apply knowledge distillation in loss but EEIL uses exemplar on the top of \cite{lwf2017}. We use exemplar without any knowledge distillation. Interestingly, FACT \cite{zhou2022forward} and Sem-aware \cite{cheraghian2021semantic} address few-shot class incremental learning in particular. Among them, Sem-aware \cite{cheraghian2021semantic} successfully applies class-semantic embedding information during training. In general, past approaches are designed aiming at 2D image data where challenges of 3D data are not addressed.
\textbf{\textit{(4)}} Our approach beats state-of-the-art methods on within and across dataset setups. We achieve superior results in each incremental task and relative accuracies. This success originates from Microshape descriptions and their ability to align with semantic prototypes, minimizing domain gaps.
\textbf{\textit{(5)}} Moreover, one can notice for all experimental setups, the performance of existing methods on base classes (1st incremental stages) are similar because of using the same PointNet features. Our approach beats them consistently. This success comes from Microshape based feature extraction. In addition, unlike other methods, our method gets benefited from language prototype and relation network.
\textbf{\textit{(6)}} Shapenet $\rightarrow$ CO3D setup has the longest sequence of tasks, arguably the most complicated of all setups. Other competing methods perform poorly in this particular setup. Our method outperforms the second best method, FACT, by up to 1.2\% accuracy in the final task, with a relative performance drop rate of 30.9\%, which is 0.4\% lower than FACT's result.

\subsection{Ablation studies}

\noindent\textbf{Effect of Microshape.} In Table \ref{table:ablation_study_cluster_w2v}, we discuss the effect of using Microshapes and (language or Microshape feature-based) prototype vectors discussed in Sec. \ref{sec:microshape_feature}. For no Microshape case, we use PointNet-like backbone features instead of Microshape-based features. We report results for different combinations of features and prototypes. One can notice that using Microshape or language prototypes individually outperforms no Microshape case. The reason is the inclusion of semantic description, transferring knowledge from base to novel classes. Also, we work on common object classes where rich language semantics are available. It could be a reason why language prototypes work better than feature prototypes. Finally, utilizing both Microshape features and language prototypes yields the best results aligning with our final recommendation.
\begin{table}[!t]
\newcolumntype{C}[1]{>{\centering\arraybackslash}p{#1}}
% \centering \small
\caption{\small Impact of \textit{\textbf{(Left)}} loss, SVD,  freezing and \textit{\textbf{(Right)}} number of centroids.
}
\scalebox{0.6}{
\begin{minipage}{0.86\linewidth}
\begin{tabular}{C{2cm}|C{0.6cm}C{0.6cm}C{0.6cm}C{0.6cm}C{0.6cm}C{0.6cm}C{0.6cm}C{0.6cm}C{0.6cm}C{0.6cm}C{0.6cm}|C{0.6cm}}
\hline
Criteria & 39 & 44 & 49 & 54 & 59 & 64 & 69 & 74 & 79 & 84 & 89 & $\Delta\downarrow$\\
\hline
No freezing & 82.3 & 75.8 & 69.4 & 65.1 & 62.7 & 61.5 & 58.3 & 57.3 & 57.9 & 52.2 & 50.7 & 38.4\\
No SVD & 82.0 & 76.7 & 72.5 & 69.7 & 67.3 & 64.5 & 61.7 & 63.0 & 61.7 & 55.2 & 54.6 & 33.4\\

$L_{mse}$ & 82.4 & 77.6 & 73.1 & 71.0 & 67.5 & 66.1 & 62.5 & 61.1 & 60.3 & 56.4 & 55.2 & 33.0\\
$L_{cross}$ & 82.6 & 77.9 & 73.9 & 72.7 & 67.7 & 66.2 & 65.4 & 63.4 & 60.6 & 58.1 & 57.1 & 30.9\\
\hline
\end{tabular}
\end{minipage} 
% \hfill 
%\hspace{0cm}
\begin{minipage}{.7\linewidth}
\centering \small
\begin{tabular}{C{0.9cm}|C{0.6cm}C{0.6cm}C{0.6cm}C{0.6cm}C{0.6cm}C{0.6cm}C{0.6cm}C{0.6cm}C{0.6cm}C{0.6cm}C{0.6cm}|C{0.6cm}}
\hline
\# & 39 & 44 & 49 & 54 & 59 & 64 & 69 & 74 & 79 & 84 & 89 & $\Delta\downarrow$\\
\hline
256 & 81.6 & 77.1 & 71.2 & 60.2 & 57.8 & 55.9 & 53.1 & 52.7 & 50.8 & 46.6 & 44.8 & 45.1\\
512 & 82.0 & 77.4 & 70.4 & 60.5 & 58.9 & 56.5 & 53.5 & 54.1 & 52.7 & 47.6 & 47.8 & 41.7\\
1024 & 82.6 & 77.9 & 73.9 & 72.7 & 67.7 & 66.2 & 65.4 & 63.4 & 60.6 & 58.1 & 57.1 & 30.9\\
2048 & 82.7 & 78 & 73.7 & 72.3 & 67.8 & 65.4 & 64.9 & 63.9 & 60.8 & 58.4 & 57.3 & 30.7\\
4096 & 82.3 & 77.2 & 72.9 & 72.0 & 67.1 & 66.1 & 64.8 & 63.0 & 60.8 & 57.9 & 57.0 & 30.7\\
\hline
\end{tabular}
\end{minipage}
}
\label{table:loss_svd_freeze}
\end{table}

\noindent\textbf{Impact of freezing, SVD and loss.} In Table \ref{table:loss_svd_freeze} \textit{\textbf{(Left)}}, We ablate our proposed method in terms of freezing, SVD, and loss. The performance gets reduced while keeping the full network trainable (no freezing) during all incremental steps. Training the full network using few-shot data during few-shot incremental steps promotes overfitting, which reduces performance. Not using the SVD step also degrades the performance because without SVD microshapes contain redundant information and do not guarantee orthogonality of microshapes. SVD step assembles the microshapes mutually exclusive and orthogonal to each other. Unlike the suggestion of \cite{8578229} regarding RelationNet architecture, we notice that the cross-entropy loss ($L_{cross}$) in Eq. \ref{eqn:total_loss_1} performs better than MSE loss ($L_{mse}$). A possible reason could be we are using RelationNet for classification, not directly to compare the relation among features and language prototypes.

\noindent\textbf{Hyperparameter sensitivity.} In Table \ref{table:loss_svd_freeze} \textit{\textbf{(Right)}}, we perform experiments varying numbers of centroids (of k-means clustering) on ShapeNet $\rightarrow$ CO3D setting. We find that a low amount of centroids (256 and 512) performs poorly in our proposed approach. However, employing 1024 (and above) centroids, the last task's accuracy and relative performance become stable. Increasing centroids up to 2048 or 4096 could not significantly differ in performance. The reason might be that applying SVD to select important centroids removes duplicate centroids, resulting in similar results for the higher number of centroids.

\subsection{Beyond FSCIL}

\begin{wraptable}{R}{0.45\textwidth}
\newcolumntype{C}[1]{>{\centering\arraybackslash}p{#1}}
\centering
\caption{\small Dynamic few-shot learning 
% results on ModelNet $\rightarrow$ ScanObjectNN
}
\scalebox{0.65}{
\begin{minipage}{0.7\textwidth}
\centering \small
\begin{tabular}{C{1.7cm}C{1.6cm}C{1.8cm}C{2cm}}
\hline
Microshape & Prototype & 1-shot & 5-shot\\
\hline
No & Feature & 39.90$\pm$0.71 & 51.65$\pm$0.35\\
Yes & Feature & 44.41$\pm$1.05 & 64.54$\pm$0.71\\
Yes & Language & \textbf{72.23$\pm$0.52} & \textbf{72.53$\pm$0.68}\\
\hline
\end{tabular}
\end{minipage}
}
\label{table:DFSL}
\end{wraptable}

% \noindent\textbf{Dynamic few-shot learning.} 

In Table \ref{table:DFSL}, we experiment on dynamic few-shot learning (DFSL), proposed by~\cite{gidaris2018dynamic}, for ModelNet $\rightarrow$ ScanObjectNN setup. DFSL is equivalent to FSCIL with only two sequences of tasks instead of various sequences in FSCIL. Note that traditionally DFSL is investigated on 2D images, but here we show results on 3D DFSL. The base classes, consisting of 26 classes, are chosen from ModelNet, but disjoint classes of the ScanObjectNN dataset, 11 classes, are used as novel classes. In each episode, the support set is built by selecting 1 or 5 examples of the novel classes, representing a 1-shot or 5-shot scenario. The query set consists of instances of the base and novel tasks. After averaging over 2000 randomly produced episodes from the test set, we evaluate few-shot classification accuracy. It has been demonstrated that utilizing more few-shot samples increases accuracy. Nevertheless, our approach (using Microshape and language prototype) can sufficiently address the DFSL problem.

\section{Conclusions}

This paper investigates FSCIL in the 3D object domain. It proposes a practical experimental setup where the base and incremental classes include synthetic and real-scanned objects, respectively. The proposed setup exhibits domain gaps related to class semantics and data distribution. To minimize the domain gap, we propose a solution to describe any 3D object with a common set of Microshapes. Each Microshape describes different aspects of 3D objects shared across base and novel classes. It helps to extract relevant features from synthetic and real-scanned objects uniquely that better aligns with class semantics. We propose new experimental protocols based on both within and cross-dataset experiments on synthetic and real object datasets. Comparing state-of-the-art methods, we show the superiority of our approach in the proposed setting.\\
\noindent\textbf{Acknowledgement:} This work was supported by North South University Conference Travel and Research Grants 2020–2021 (Grant ID: CTRG-20/SEPS/04).

\clearpage
{
\small
\bibliographystyle{splncs04}
\bibliography{egbib}
}

\title{Supplementary material for\\``Few-shot Class-incremental Learning for\\3D Point Cloud Objects"} % Replace with your title

\begin{comment}
\titlerunning{ECCV-22 submission ID \ECCVSubNumber} 
\authorrunning{ECCV-22 submission ID \ECCVSubNumber} 
\author{Anonymous ECCV submission}
\institute{Paper ID \ECCVSubNumber}
\end{comment}
%******************

% \begin{comment}
\titlerunning{FSCIL for 3D Point Cloud Objects}

\author{Townim Chowdhury\inst{1}\orcid{0000-0003-1780-6046} \and
Ali Cheraghian\inst{2,3}\orcid{0000-0002-3324-7849} \and
Sameera Ramasinghe\inst{4}\orcid{0000-0002-3200-9291} \and\\
Sahar Ahmadi\inst{5}\orcid{0000-0002-2161-7005} \and
Morteza Saberi\inst{6}\orcid{0000-0002-5168-2078} \and
Shafin Rahman\inst{1}\thanks{Corresponding author}\orcid{0000-0001-7169-0318}
}
\authorrunning{T. Chowdhury et al.}

\institute{Dept. of Electrical and Computer Engineering, North South University, Bangladesh \and
School of Engineering, Australian National University, Australia \and
Data61, Commonwealth Scientific and Industrial Research Organisation, Australia
\and
Australian Institute for Machine Learning, University of Adelaide, Australia 
\and
Business school, The University of New South Wales,  Australia
\and
School of Computer Science and DSI, University of Technology Sydney, Australia
\\
% Corresponding Author Email: \email{\small\texttt{shafin.rahman@northsouth.edu}}
\email{\small\texttt{\{townim.faisal, shafin.rahman\}@northsouth.edu,\\ali.cheraghian@anu.edu.au,  sameera.ramasinghe@adelaide.edu.au,\\sahar.ahmadi@unsw.edu.au, morteza.saberi@uts.edu.au}}
}

% \end{comment}
%******************
\maketitle

\begin{abstract}
This supplementary material provides additional details in support of the contribution presented in the main paper.
\begin{itemize}
\item[$\bullet$] Section \ref{arch_details}: Architecture Details (additional discussion in support of Section 3.1 of the main paper). 
\item[$\bullet$] Section \ref{k_means}: K-Means Visualization (additional discussion in support of Section 3.2 of the main paper).
\item[$\bullet$] Section \ref{exp_setup}: Details of Experimental Setup (additional discussion in support of Section 4 of the main paper).
\item[$\bullet$]  Section \ref{beyond_fscil}: 3D Object Recognition (additional discussion in support of Section 4.3 of the main paper).
\end{itemize}

\end{abstract}

\section{Architecture Details}
\label{arch_details}

Here, we provide more details on the backbone network, $F$ and the relation module, $R$.

\begin{figure} [!h]
%\centering
\includegraphics[width=1\linewidth,trim=0cm 0cm 0cm 0cm, clip]{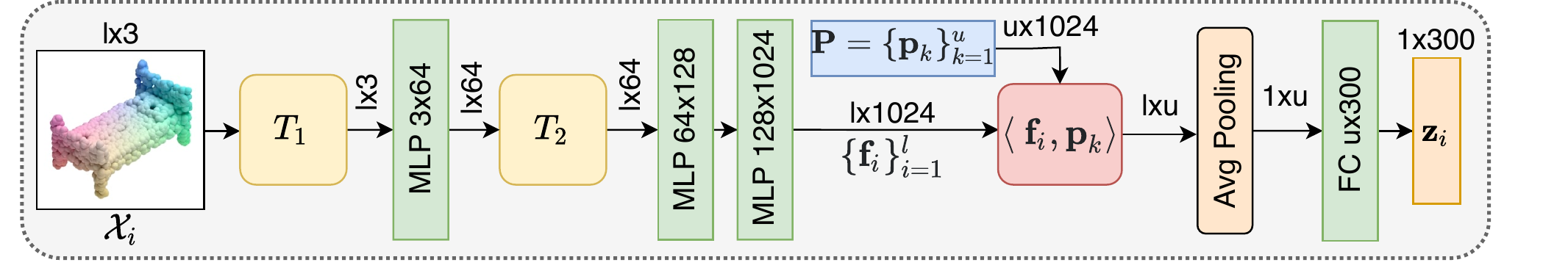}
% left lower right up
%\hspace{2em}
\caption{\small Detailed network architecture the backbone. $T_1$ and $T_2$ are transformation networks for inputs of $l$ points
}
\label{fig:projection_network}
% \vspace{-1cm}
\end{figure}

\noindent\textbf{Backbone network.} A mini transformation network $T_1$ \cite{jaderberg2015spatial} takes raw input point clouds of $n$ points, and the output passes into a shared multi-layer perceptron network of output size 64. This output matrix passes into another feature transformation network $T_2$, and the transformed output matrix then passes into a shared multi-layer perceptron network with layer output sizes 64, 128, 1024. This extracted features of $n$ points, $\textbf{f}_i$ is forwarded with Microshape basis, $\textbf{p}_{k}$ in a inner product function, $\langle\ \textbf{f}_{i},\textbf{p}_{k}\rangle$. Then, we calculate the average similarity vector from the output with average pooling. The similarity vector is passed into a fully connected layer (of 300 dimensions), generating the features $\textbf{z}_i$. Here, all layers include ReLU and batch normalization.

\begin{figure} [!t]
\centering
\includegraphics[width=1\linewidth,trim=0cm 0cm 0cm 0cm, clip]{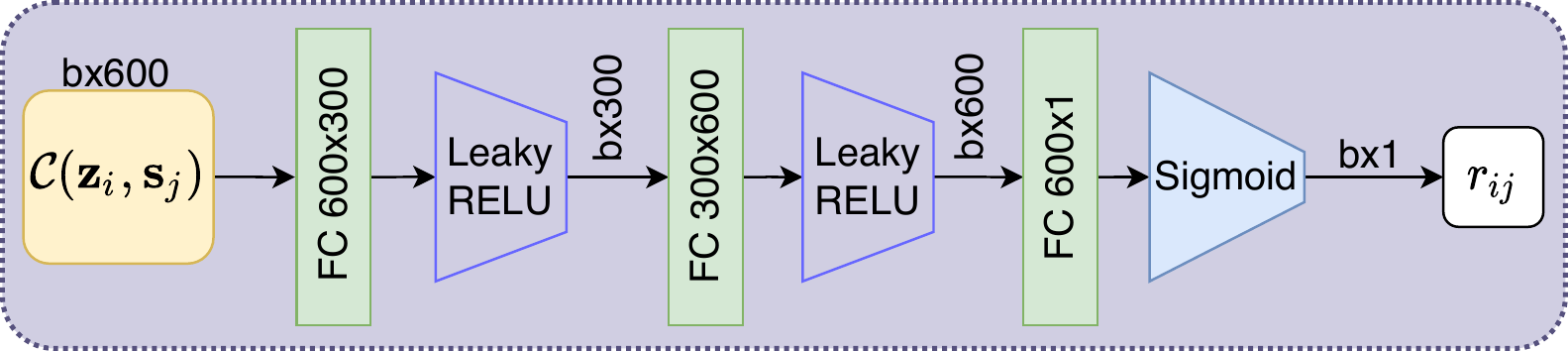}
% left lower right up
%\hspace{2em}
\caption{\small Detailed architecture for Relation module $R$. `b' means batch size and $r_{ij}$ is the relation score.}
\label{fig:relation_network}
% \vspace{-1cm}
\end{figure}

\noindent\textbf{Relation Module architecture.} The input of relation module is generated by a concatenation function $\mathcal{C}(\textbf{z}_{i},\textbf{s}_{j})$, that takes feature of point cloud object, $\textbf{z}_i$ and the semantic embedding of the task $\textbf{s}_j$. This generated input is passed into three fully-connected layers (300,600,1). Except for the output layer, which is Sigmoid and generates relation scores, $r_{ij}$, all fully-connected layers are associated with LeakyReLU.

\begin{figure}[!t]
\centering
% trim={<left> <lower> <right> <upper>}
\includegraphics[width=1\textwidth,trim={0.2cm 0.2cm 0.2cm 0.2cm},clip]{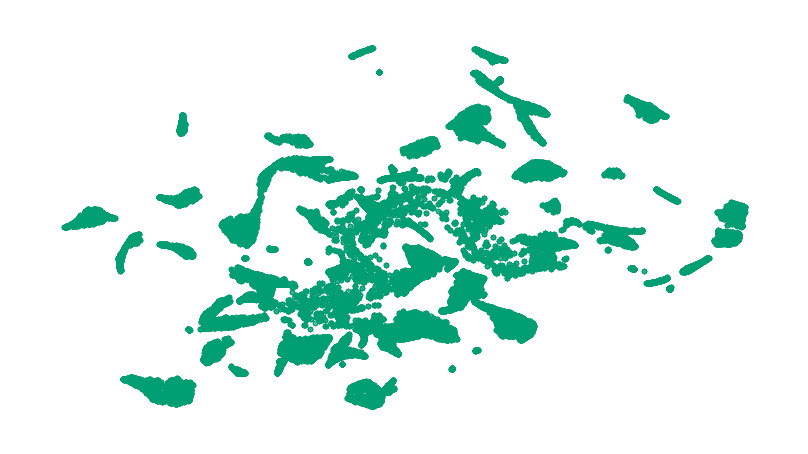}
% \vspace{-1em}
\caption{tSNE visualization for points in different 3D point cloud objects}
%\vspace{-1em}
\label{fig:tsne-points}
\end{figure}

\section{K-Means Visualization}
\label{k_means}

We plot a tSNE visualization for all points in 10000 random different 3D point cloud objects in Fig. \ref{fig:tsne-points}. We notice that some clusters have been formed from where we calculated the Microshapes.

\section{Details of Experimental Setup}
\label{exp_setup}

We experiment on two synthetic datasets i.e. ModelNet40 \cite{modelnet2015}, ShapeNet \cite{shapenet15} and three real-scanned dataset i.e. ScanObjectNN \cite{scanobjectnn19}, Common Objects in 3D (CO3D) \cite{co3d21} with our proposed two different experimental setups.

\subsection{Within-dataset Experiment}

\begin{figure}[htbp]
 \begin{minipage}{1\linewidth}
  \centering
  \caption*{(a) ModelNet}
  \includegraphics[width=1\textwidth,trim={0cm 0cm 0cm 0cm},clip]{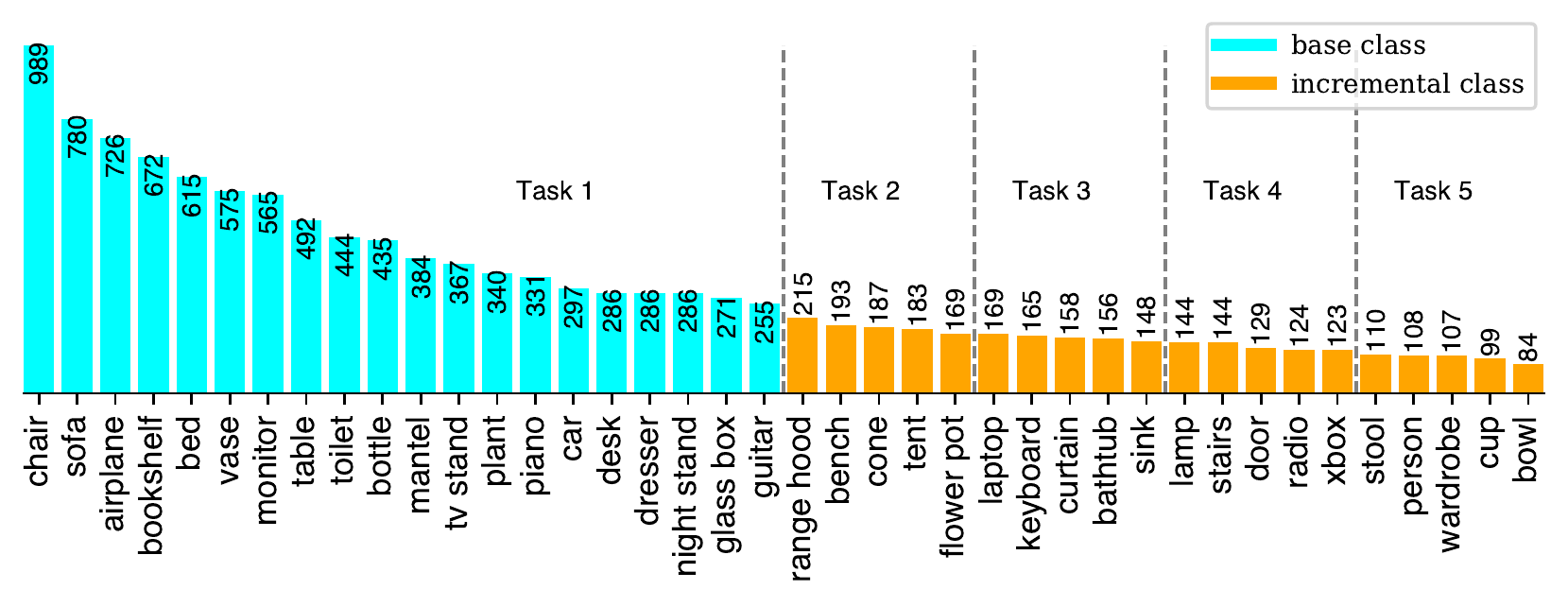}
  \label{fig:m40_dataset}
 \end{minipage}%
 
 \begin{minipage}{1\linewidth}
  \centering
  \caption*{(b) ShapeNet}
  \includegraphics[width=\textwidth,trim={0cm 0cm 0cm 0cm},clip]{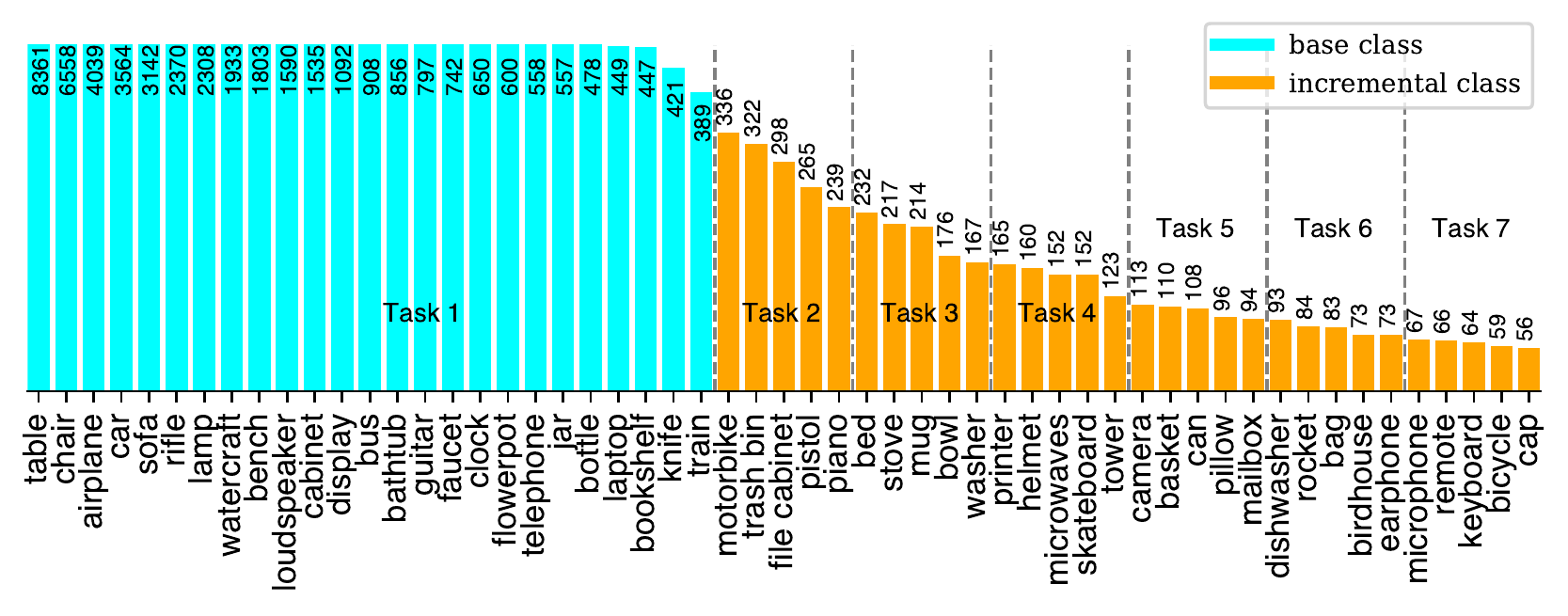}
  \label{fig:shapenet_dataset}
 \end{minipage}
 
 \begin{minipage}{1\linewidth}
  \centering
  \caption*{(c) CO3D}
  \includegraphics[width=\textwidth,trim={0cm 0cm 0cm 0cm},clip]{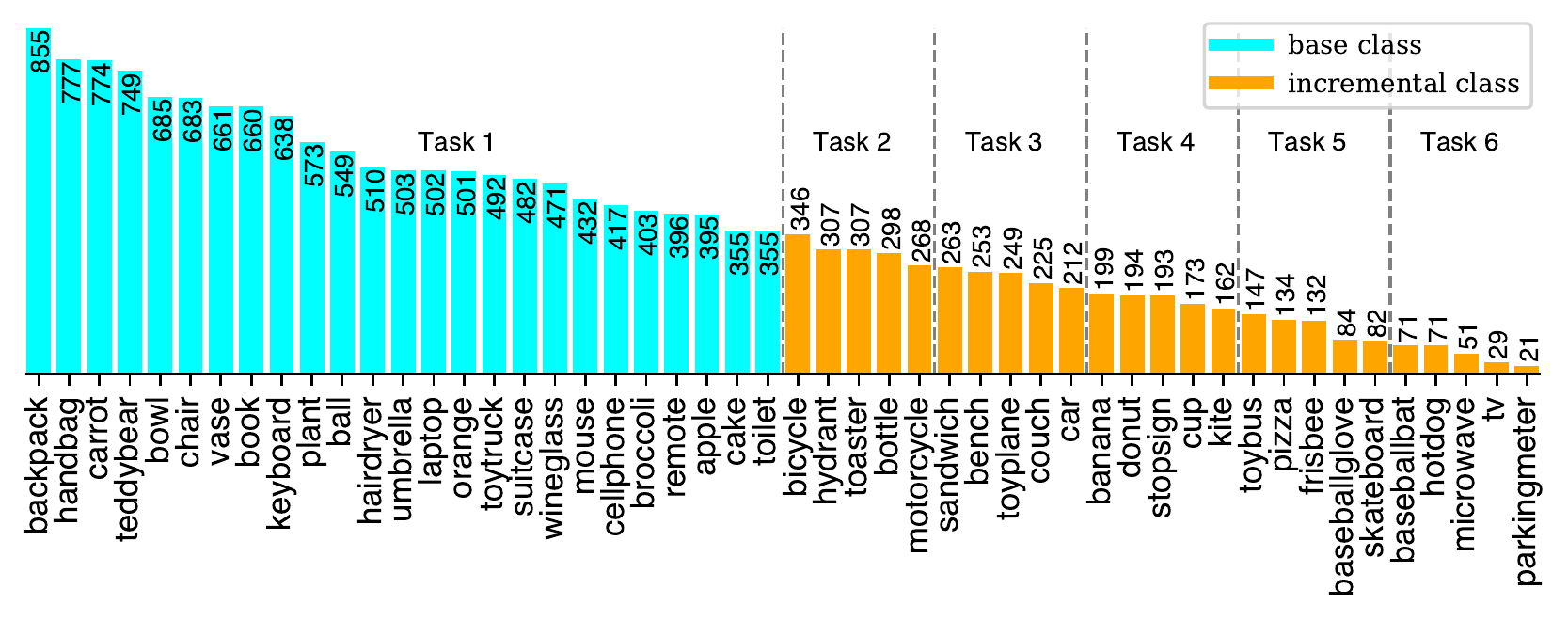}
  \label{fig:co3d_dataset}
 \end{minipage}
 \caption{\small Data distribution for within-dataset experiments by sorting all classes in descending order based on instance frequency. The plot clearly shows that all three setups follow a long-tail distribution. In the experimental setup of (b) ShapeNet, some long bars have been clipped for better visualization.}
 \label{fig:within_dataset}
\end{figure}

We design within-dataset experimental setups by ordering all classes in descending order for a dataset based on sample frequency. It assists us in distinguishing between base and novel classes since base classes have more instances than novel classes. Rare objects have fewer samples in the actual world. So, following this order, we create a realistic experimental setting. Therefore, all within-dataset experiments follow long-tail distribution, shown in Fig. \ref{fig:within_dataset}. The base and incremental classes are treated as the head and tail classes of this data distribution, respectively. \\
\textit{\textbf{(1)} ModelNet40:} It comprises 12,311 3D point cloud objects from 40 categories. We select 20 classes as base classes with 7438 training instances and 1958 test instances. The rest of the 20 classes are used for four incremental tasks consisting of 510 test instances. \\
\textit{\textbf{(2)} ShapeNet:} It has 50604 shapes from 55 categories. We select 25 classes as base classes with the topmost training instances and a total of 36791 training and 9356 test samples. Then we choose the rest of the 30 classes as few-shot incremental classes with 887 test instances. \\
\textit{\textbf{(3)} CO3D:} It is composed of 50 MS-COCO types of 3D point clouds. We choose 25 base classes, with 12493 training and 1325 test instances. The remaining 25 classes with 407 test instances are utilized for incremental training. \\

\subsection{Cross-dataset Experiment}
For cross-dataset experiments, we choose synthetic dataset as base class and real-scanned dataset as novel class. Table \ref{table:cross_dataset} shows the detailed data distribution for three experimental setups.\\
\textit{\textbf{(1)} ModelNet40 $\rightarrow$ ScanObjectNN:} We follow the selection of classes from \cite{cheraghian2021zero}. Here, we select 26 base classes from ModelNet40. On the other hand, ScanObjectNN has 15 classes with 2902 3D point cloud objects, but we choose non-overlapped 11 classes from ScanObjectNN for incremental tasks as novel classes.\\
\textit{\textbf{(2)} ShapeNet $\rightarrow$ ScanObjectNN:} We select 44 disjoint classes from ShapeNet as base classes. For base model training, there are 22318 training and 5845 test instances from ShapeNet. Also, from ScanObjectNN, we select all 15 classes for few-shot incremental model training with 581 test instances.\\
\textit{\textbf{(3)} ShapeNet $\rightarrow$ CO3D:} Base classes are selected from Shapenet, while for few-shot incremental steps, the classes are chosen from CO3D. This setup represents the most realistic scenario. We select non-overlapped 39 classes from Shapenet for base classes. Also, CO3D has 50 classes with 16557 training and 1732 test instances. Here, we choose all 50 classes for few-shot incremental tasks. \\

\begin{table}[!t]
% \centering \small
\newcolumntype{C}[1]{>{\centering\arraybackslash}p{#1}}
\caption{\small Details of cross-dataset experimental setup. Task 1 represents the base class, whereas the rest of the tasks represent the novel class}
\scalebox{0.6}{
\begin{minipage}{0.73\textwidth}
% \caption*{ShapeNet }
\begin{tabular}{C{1.4cm}|C{0.7cm}|C{6cm}}
\hline
\multicolumn{3}{c}{\textbf{(a) ShapeNet $\rightarrow$ CO3D}}\\
\hline
Dataset & Task & Name of class \\
\hline 
ShapeNet & 1 & airplane, trash bin, basket, bathtub, bed, birdhouse, bookshelf, bus, cabinet, camera, can, cap, clock, dishwasher, display, faucet, file cabinet, guitar, helmet, jar, knife, lamp, loudspeaker, mailbox, microphone, mug, piano, pillow, pistol, flowerpot, printer, rifle, rocket, stove, table, tower, train, vessel, washer\\
\hline
\multirow{10}{*}{CO3D} & 2 & apple, backpack, ball, banana, baseballbat\\
& 3 & baseballglove, bench, bicycle, book, bottle\\
& 4 & bowl, broccoli, cake, car, carrot\\
& 5 & cellphone, chair, couch, cup, donut\\
& 6 & frisbee, hairdryer, handbag, hotdog, hydrant\\
& 7 & keyboard, kite, laptop, microwave, motorcycle\\
& 8 & mouse, orange, parking meter, pizza, plant\\
& 9 & remote, sandwich, skateboard, stopsign, suitcase\\
& 10 & teddybear, toaster, toilet, toybus, toyplane\\
& 11 & toytruck, tv, umbrella, vase, wineglass\\
\hline
\end{tabular}
\end{minipage} \hfill
\begin{minipage}{0.53\textwidth}
\begin{tabular}{C{2cm}|C{0.7cm}|C{8cm}}
\hline
\multicolumn{3}{c}{\textbf{(b) ShapeNet $\rightarrow$ ScanObjectNN}}\\
\hline
Dataset & Task & Name of class \\
\hline
ShapeNet & 1 & airplane, basket, bathtub, bench, bicycle, birdhouse, bottle, bowl, bus, camera, can, cap, car, clock, keyboard, dishwasher, earphone, faucet, file cabinet, guitar, helmet, jar, knife, lamp, laptop, loudspeaker, microphone, microwaves, motorbike, mug, piano, pistol, flowerpot, printer, remote, rifle, rocket, skateboard, stove, telephone, tower, train, watercraft, washer\\
\hline
\multirow{3}{*}{ScanObjectNN} & 2 & bag, bin, box, cabinet, chair\\
& 3 & desk, display, door, shelf, table\\
& 4 & bed, pillow, sink, sofa, toilet\\
\hline
\hline
\multicolumn{3}{c}{\textbf{(c) ModelNet $\rightarrow$ ScanObjectNN}}\\
\hline
Dataset & Task & Name of class \\
\hline
ModelNet & 1 & airplane, bathtub, bottle, bowl, car, cone, cup, curtain, flower pot, glass box, guitar, keyboard, lamp, laptop, mantel, night stand, person, piano, plant, radio, range hood, stairs, tent, tv stand, vase\\
\hline
 \multirow{3}{*}{ScanObjectNN} & 2 & cabinet, chair, desk, display\\
& 3 & door, shelf, table, bed\\
& 4 & sink, sofa, toilet\\
\hline
\end{tabular}
\end{minipage}
}
\label{table:cross_dataset}
\end{table}

\begin{table}[h]
\newcolumntype{C}[1]{>{\centering\arraybackslash}p{#1}}
\caption{\small 3D recognition on common objects of ModelNet40 (synthetic training) and ScanObjectNN (real-scanned testing)}
\centering \small
\begin{tabular}{C{6cm}C{2cm}}
\hline
Method & Accuracy \\
\hline
Baseline (without Microshape) & 44.25\\
Ours (with Microshape) & \textbf{46.34}\\
\hline
\end{tabular}
\label{table:object_recognition}
\end{table}

\begin{figure*}[t]
  \centering
    \includegraphics[width=1\textwidth,trim={0cm 0cm 0cm 0cm},clip]{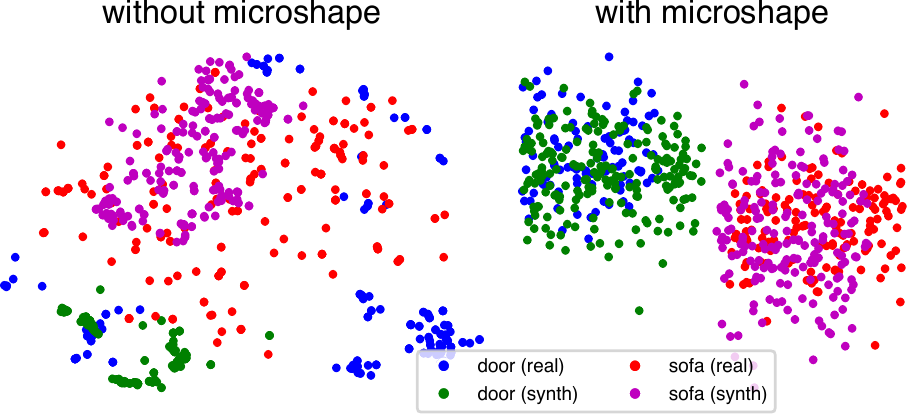}
  \caption{\small Effect of using Microshape. Only two classes, `door' and `sofa', are used for visualization. Synthetic and real instances form different clusters on the left because of the domain gap. Microshape-based feature minimizes this gap by mixing both instances in the same cluster}
  \label{fig:domaingap}
\end{figure*}

\section{3D Object Recognition}
\label{beyond_fscil}
Microshape based 3D point description can benefit many problems beyond FSCIL.%\\
% \noindent\textbf{3D object recognition.} 
In Table \ref{table:object_recognition}, we perform 3D object recognition experiment on 11 common objects of ModelNet40 and ScanObjectNN. Here, the model is trained on the common (synthetic) objects of ModelNet40 and evaluated on the same (real-scanned) objects from ScanObjectNN. 
There is a clear domain gap from the data distribution of training (synthetic) and testing (real-scanned) instances. We attempt to reduce this gap using our Microshape based backbone and relation network (see Fig. \ref{fig:domaingap}). 
Our approach combining of Microshape and relation network with language prototype outperforms the baseline (without Microshape). It is possible because our method helped reducing the domain gap of training and testing data.

\clearpage

\end{document}